\documentclass[letterpaper, 10 pt, conference]{ieeeconf} 

\IEEEoverridecommandlockouts 

\usepackage[pdftex]{graphicx}
\usepackage{amsmath} 
\usepackage{amssymb} 
\usepackage{comment}
\usepackage{cite}

\usepackage[switch]{lineno}
\newcommand*\patchAmsMathEnvironmentForLineno[1]{%
 \expandafter\let\csname old#1\expandafter\endcsname\csname #1\endcsname
 \expandafter\let\csname oldend#1\expandafter\endcsname\csname end#1\endcsname
 \renewenvironment{#1}%
 {\linenomath\csname old#1\endcsname}%
 {\csname oldend#1\endcsname\endlinenomath}}%
\newcommand*\patchBothAmsMathEnvironmentsForLineno[1]{%
 \patchAmsMathEnvironmentForLineno{#1}%
 \patchAmsMathEnvironmentForLineno{#1*}}%
\AtBeginDocument{%
\patchBothAmsMathEnvironmentsForLineno{equation}%
\patchBothAmsMathEnvironmentsForLineno{align}%
\patchBothAmsMathEnvironmentsForLineno{flalign}%
\patchBothAmsMathEnvironmentsForLineno{alignat}%
\patchBothAmsMathEnvironmentsForLineno{gather}%
\patchBothAmsMathEnvironmentsForLineno{multline}%
}

\graphicspath{{./imgs/}}
\newcommand{\eqrf}[1]{Eq.\ (\ref{#1})}
\newcommand{\firf}[1]{Fig.\ \ref{#1}}
\newcommand{\tarf}[1]{Table \ref{#1}}
\newcommand{\sref}[1]{Section \ref{#1}}

\title{\LARGE \bf
Emotional Speech Synthesis for Companion Robot to Imitate \\
Professional Caregiver Speech 
}
\author{
	\authorblockN{Takeshi Homma$^{1}$ \,\, Qinghua Sun$^{1}$ \,\, Takuya Fujioka$^{1}$ \,\, Ryuta Takawaki$^{2}$ \,\, Eriko Ankyu$^{2}$ \\
	Kenji Nagamatsu$^{1}$ \,\, Daichi Sugawara$^{2}$ \,\, Etsuko T. Harada$^{2}$%
	}
\thanks{$^{1}$T. Homma, Q. Sun, T. Fujioka, and K. Nagamatsu are with R\&D Group, Hitachi, Ltd., 1-280 Higashi-koigakubo, Kokubunji, Tokyo 185-8601, JAPAN.
	{\tt\small takeshi.homma.ps@hitachi.com}}
\thanks{$^{2}$R. Takawaki, E. Ankyu, D. Sugawara, and E.T. Harada are with Faculty of Human Science, University of Tsukuba, 1-1-1 Tennodai, Tsukuba, Ibaraki 305-8577, JAPAN.}%
\thanks{This study was approved by the IRB in Faculty of Human Science, University of Tsukuba on February 7, 2020.}
}

\begin{document}

\maketitle
\thispagestyle{empty}
\pagestyle{empty}

\begin{abstract} 
When people try to influence others to do something, they subconsciously adjust their speech to include appropriate emotional information. In order for a robot to influence people in the same way, the robot should be able to imitate the range of human emotions when speaking. To achieve this, we propose a speech synthesis method for imitating the emotional states in human speech. In contrast to previous methods, the advantage of our method is that it requires less manual effort to adjust the emotion of the synthesized speech. Our synthesizer receives an emotion vector to characterize the emotion of synthesized speech. The vector is automatically obtained from human utterances by using a speech emotion recognizer. We evaluated our method in a scenario when a robot tries to regulate an elderly person's circadian rhythm by speaking to the person using appropriate emotional states. For the target speech to imitate, we collected utterances from professional caregivers when they speak to elderly people at different times of the day. Then we conducted a subjective evaluation where the elderly participants listened to the speech samples generated by our method. The results showed that listening to the samples made the participants feel more active in the early morning and calmer in the middle of the night. This suggests that the robot may be able to adjust the participants' circadian rhythm and that the robot can potentially exert influence similarly to a person.
\end{abstract}

\section{Introduction}
To improve the quality of life of the elderly, companion robots with communication capabilities are being introduced into the field of elderly care\cite{Inoue,IR}. The robots provide elderly users with appropriate suggestions to encourage them to do spontaneous activities to maintain their health. The challenge here is whether the user will follow suggestions from a robot rather than a person. Specifically, we aim to clarify whether the user can exchange emotional states and empathize with the robot and follow its suggestions.

Such a challenge is not always successful even in human conversations. In human conversations, there is a constant emotional exchange. To convince someone to listen to a suggestion, the speaker subconsciously alters their speech to include appropriate emotional information to influence the listener. To make human-robot conversations more natural, robots should be able to reproduce such emotional control in their synthesized speech.

Many methods for emotional speech synthesis have been proposed \cite{Yamagishi, Lorenzo-Trueba, Xue, Rabiee, Zhu, Lorenzo-Trueba2, Nose, Hodari, Cai}. However, the main drawback of these methods is that they require the developers to adjust the emotional parameters until the synthesizer generates the desired speech samples that reproduce the emotion in human speech. It is difficult for the developers to imagine the relationship between the emotional parameters and the voice characteristics to be generated. Therefore, this process must be carried out in a trial-and-error manner and requires much effort.

In this study, we propose a speech synthesis method to reproduce the emotions that appear in the voices of target speakers while minimizing the need to manually adjust the emotion. The target speakers refer to the professional caregivers who are performing the conversational behavior that the robot aims to reproduce. In our speech synthesis method, emotional parameters, which are numerical values characterizing the emotional state of speech, are used as input features in training the synthesis model. During speech synthesis, the emotion parameter estimated from the target speaker's voice is input to the synthesizer so that the synthesizer can generate speech along with the target speaker's emotion. The emotional parameters are automatically estimated by an emotion recognizer trained separately. This way, our method enables the generation of emotional speech and minimizes the need to manually adjust the emotional parameters.

We evaluate the effectiveness of our speech synthesis method in situations where a robot tries to influence people by speaking in emotional voices. In our experimental conditions, we assume the robot is in an elderly residence and offers suggestions to help adjust the elderly person's circadian rhythm. Studies have shown that as people age, their circadian rhythm becomes disrupted \cite{Hood}, and a large percentage of elderly who live alone have sleep disorders \cite{Livingston}. We designed our experiment to address this issue. To help regulate their rhythm, we aim to increase their arousal level in the early morning and decrease it in the middle of the night through utterances from the robot. For the target speakers, we recruited professional caregivers working in residential facilities for the elderly because they frequently engage with elderly people in daily conversation and provide encouragement.

\section{Related Work}
Previous studies on emotional speech synthesis assumed that emotions were divided into categories \cite{Yamagishi, Lorenzo-Trueba} or specified by a multidimensional parameter where each dimension corresponds to an emotional aspect \cite{Xue, Rabiee, Zhu}. With the aim of creating a synthesizer that generates voices with emotional information that matches the listener's perceived emotion, some studies \cite{Lorenzo-Trueba2, Nose} utilized emotional parameters based on the listener's perception instead of the speaker's intention to characterize the emotional aspects of speech. However, the emotional parameters for synthesis needed to be adjusted by developers.

In addition to the speech synthesizer-based approach above, voice conversion technology is also utilized for emotional speech generation. This approach assumes the existence of two speech data, an original voice and a target voice. The emotional aspect of the original voice is converted to align with that of the target voice while the speaker identity of the original voice remains unchanged\cite{Shankar, Zhou, Li}. However, this approach is limited in terms of the quality of emotional naturalness and stability of the generated voices. For a robot to be capable of emotional speech, the quality of its voices must be stable. Therefore, the speech synthesizer-based approach is more ideal for practical use than the voice conversion-based approach thus far.

In the synthesizer-based approach, to minimize the effort needed to annotate the emotional information in the training corpus, several studies \cite{Hodari, Cai} have proposed methods where each speech sample in the corpus is labeled with emotional parameters fed by a separately trained emotion recognizer. Our study similarly utilizes the emotional parameters obtained by the recognizer to construct the training corpus. However, we also incorporate the emotional parameters obtained from the speech of the target speakers to specify emotional information in the synthesis process. Because the emotional parameters for synthesis are automatically obtained without manual adjustment, it drastically reduces the effort required on the developer's part to adjust the parameters to generate desirable speech.

Studies have shown that controlling the emotional aspects of the voices of robots and agents is beneficial to users in various applications, e.g., medication reminders \cite{James}, learning support agents \cite{Dmello}, non-task oriented dialogue systems \cite{Chiba}, and in-vehicle agents \cite{Nass}. To the best of our knowledge, our study is the first to show that a robot with emotional speech synthesis may be able to help regulate a person's circadian rhythm. Building on the previous emotional speech synthesis methods, our method makes it possible for a synthesized voice to imitate the emotional state of a target human voice.

\section{Proposed Pipeline}
\begin{figure}
	\centering
	\includegraphics[width=0.97\linewidth]{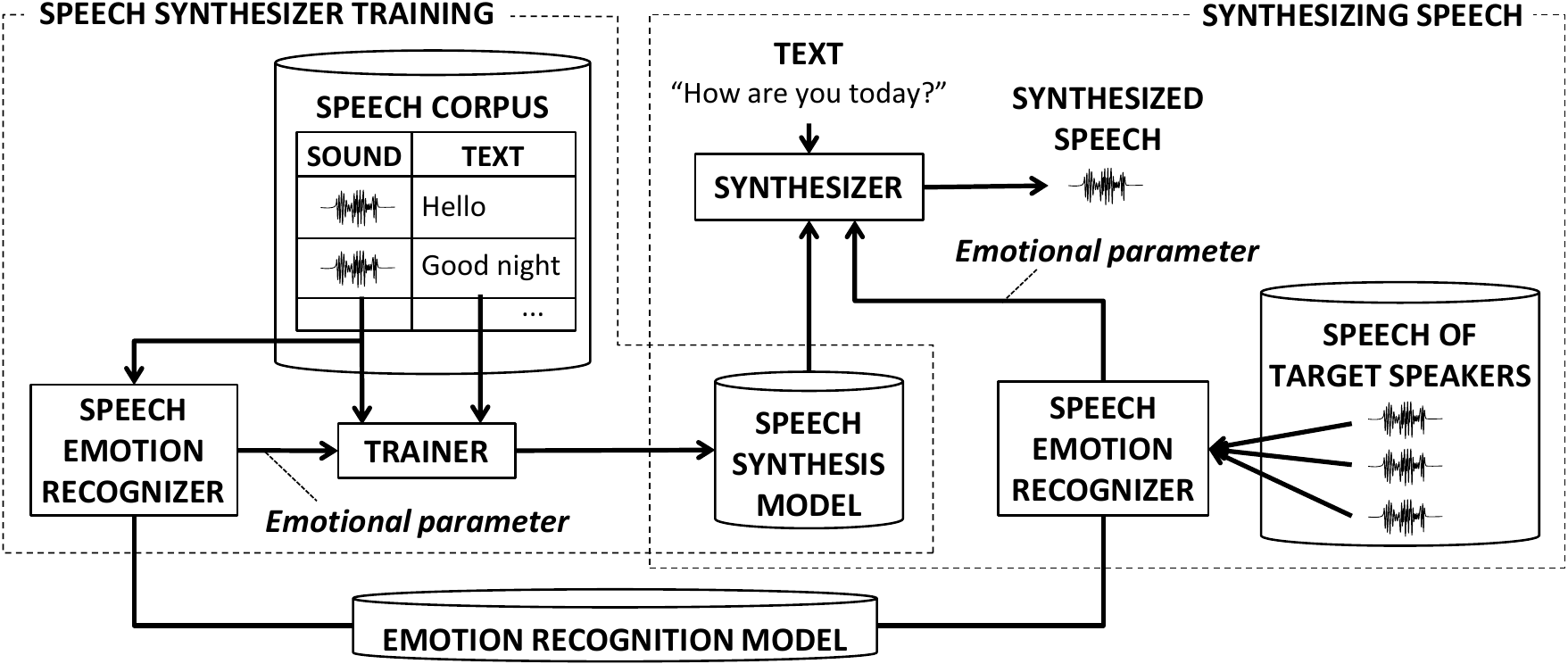}
	\caption{Pipeline of emotional speech synthesizer.}
	\label{fig:pipeline}
\end{figure}

To reproduce the human use of emotions in speech synthesis, we propose a method that combines speech synthesis with emotional speech recognition. \firf{fig:pipeline} shows the pipeline of the proposed method.

\subsection{Training of Emotional Speech Synthesis Model}
Speech synthesizers typically receive linguistic features derived from text as an input feature and generate a speech waveform as output. We add an emotion vector to the input feature to specify the emotion of the speech. An emotion vector is a multidimensional vector in which each dimension shows the intensity of the corresponding emotional aspect. To train a speech synthesizer model, we first prepare a speech corpus recorded by one voice actor. The corpus contains many speech data and corresponding text. Using the corpus, we obtain the emotion vector of each speech sample by inputting it to a speech emotion recognizer. In training, we use both linguistic features and emotion vectors as input features. In the speech synthesis process, we input the linguistic features of a target text with an emotion vector that specifies the emotion of the generated speech.

\subsection{Speech Synthesis Conditioned by Emotion of Target Speech}
The emotion vector for speech synthesis is determined on the basis of the target speaker's voice.

\subsubsection{Emotional Statistics Calculation of Target Speech}
We denote a target speaker as $t$ ($t \in T=\{t_1, t_2, \dots, t_{|T|}\}$). We assume that each target speaker speaks under multiple conditions. Let the condition be $c$ ($c \in C=\{c_1, c_2, \dots, c_{|C|}\}$). We assume the target speakers determine the emotion in their voices depending on the condition. In a later section, we will  discuss our investigation on how professional caregivers change the emotions in their voices. We studied how the caregivers change the emotion in voices depending on the time of day and the elderly person's utterances to which they are responding. Thus, the experimental target speaker conditions include the time of day and the elderly person's utterances.

An utterance spoken by speaker $t$ under condition $c$ is denoted as $x_{t,c}$. We refer to this as the target speech. From the emotion vectors of target utterance in a single condition, we calculate the averaged emotion vector $\boldsymbol{e}_{c}$ for each condition $c$ as follows.

\begin{align}
	\boldsymbol{e}_{c} = \frac{1}{|T|} \sum_{t \in T} {\rm SER}(x_{t,c}) \label{eqn:1}
\end{align}
${\rm SER}$ is a function for obtaining an emotion vector from a given utterance using the speech emotion recognizer. The function outputs an $E$-dimensional emotion vector.

\subsubsection{Emotional Input for Speech Synthesis}
It should be possible to generate an utterance with the same emotion as that of the target speech by simply inputting the calculated emotion vector to the synthesizer. However, our preliminary experiment showed that using the calculated vectors does not always produce speech with the same emotion as that of the target speech because of differences in the emotion distributions. These differences exist between the speech synthesis corpus and the target speech and between the emotional variation of averaged vectors across conditions and the variation within a single target speaker. Thus, we transformed the distribution of the target speech's emotion vectors. Specifically, we set the average of the emotion for speech synthesis to be the same as that of the speech synthesis corpus, and the variance of the emotion to be the same as that of a single target speaker. The average emotion vector in the speech synthesizer corpus $\boldsymbol{a}$ is obtained as follows.

\begin{equation}
	\boldsymbol{a} = \frac{1}{|W|} \sum_{w \in W} {\rm SER}(w)
\end{equation}
Each utterance in the corpus is denoted as $w \in W=\{w_1, w_2, \dots, w_{|W|}\}$. The emotion recognizer (${\rm SER}$) is the same as the one in \eqrf{eqn:1}. The emotion vector to be input to the synthesizer $\boldsymbol{e}'_{c}$ is calculated by converting $\boldsymbol{e}_{c}$ in the following equation.
\begin{align}
	\boldsymbol{e}'_{c} = \alpha \cdot \lambda \left( \boldsymbol{e}_{c} - \boldsymbol{e} \right) + \boldsymbol{a}
	\label{eqn:conv}
\end{align}
$\boldsymbol{e}$ is the average emotion vector in the target speech obtained as follows.
\begin{align}
	\boldsymbol{e} = \frac{1}{|C|} \sum_{c \in C} {\boldsymbol{e}_{c}}
\end{align}

The coefficients $\alpha$ and $\lambda$ adjust emotional variations. To reproduce the variations produced by an individual target speaker, we calculate the coefficient $\lambda$ to adjust the variance across conditions using the following equation.
\begin{align}
V_t &= \frac{1}{|T|} \sum_{t \in T} {\rm Var}_{c} \left( {\rm SER}(x_{t,c}) \right), \,\,\, V_e = {\rm Var}_{c} (\boldsymbol{e}_{c}) \\
\lambda &= \sqrt{ \left( \Sigma_{i=1}^{E} V_t^i \right) / \left( \Sigma_{i=1}^{E} V_e^i \right) }
\end{align}
${\rm Var}_c$ denotes the variance of the vectors across conditions $c$. $V_t^i$ and $V_e^i$ denote the $i$-th element of vectors $V_t$ and $V_e$, respectively. Theoretically, the variation of emotional parameters should be different between emotional dimensions so that the coefficient $\lambda$ can be calculated for each emotional dimension separately. However, when there are few conditions, the coefficient separately calculated for each dimension becomes unreliable. For example, the variation may become extremely small. To calculate a reliable coefficient, we use a single coefficient taking all the emotional dimensions into account.

The coefficient $\alpha$ in \eqrf{eqn:conv} is the only parameter adjusted by developers to generate voices with the desired emotions.

\section{Pipeline Implementation}
\subsection{Speech Emotion Recognizer}
\label{ss:emotion_recognizer}
\begin{table}[tb]
	\caption{Six emotional dimensions evaluated by automatic speech emotion recognizer.}
	\label{tbl:6axis}
	\centering
	{\footnotesize
		\begin{tabular}{ll}
			\hline
			Name & Scoring Adjective (Min.1 -- Max.7) \\
			\hline
			Pleasantness & Unpleasant - Pleasant \\
			Arousal & Sleepiness - Arousal \\ 
			Dominance & Obedience - Dominance \\
			Credibility & Distrust - Trust \\
			Interest & Indifference - Interest \\
			Positivity & Negation - Affirmation \\ 
			\hline
		\end{tabular}
	}
\end{table}

In the majority of previous studies on speech emotion recognition, the number of emotional dimensions range from two to four \cite{Karadogan,Eyben,Yang,Schmitt,Atmaja,Grimm,Parthasarathy,Moore,Schuller}. In our study, to characterize speech from a wider range of emotional aspects, we implemented the 6-dimensional emotions proposed by Mori et al.\cite{MoriAST} (\tarf{tbl:6axis}) They also published an emotional speech corpus called UUDB\cite{1400Mori}, which contains Japanese speech data and their corresponding emotion labels based on the 6-dimensional emotions. We trained an emotion recognizer on the basis of UUDB. For the emotion recognition algorithm, we used a deep learning-based method proposed by Fujioka et al.\cite{Fujioka} which takes labeling ambiguity into account. In UUDB, three labelers scored each speech sample in accordance with the definitions of the six dimensions.

We trained the recognition model using the average of the three labelers' scores as the ground-truth. Each dimension of the output vector from the recognizer is a continuous number from 1 to 7. The average prediction error for UUDB was 0.490, which is lower than the error (0.526) reported by Mori et al. \cite{1400Mori}

\subsection{Speech Synthesizer}
\subsubsection{Voice Actor Recording}
We collected utterances from one female voice actor to build the speech corpus for training the synthesizer. To enable the synthesizer to generate speech with various emotions, we asked the actor to say sentences with 55 types of emotions following the JEITA guidelines for emotions expressed in speech \cite{JEITAKikaku}. We also recorded utterances with three additional emotions: happiness, anger, and sadness. We recorded whispering voices too. For the emotion in the sentences to sound natural, the actual words in the sentences must not contradict the emotion. Thus, we designed the sentences such that they sounded natural for the emotion that they were meant to express. We recorded 12 hours of speech data containing 6.7k sentences in total, and by using the speech emotion recognizer described in \sref{ss:emotion_recognizer}, we obtained 6-dimensional emotion vectors from the speech.

\subsubsection{Synthesizer Model Training}
We built the speech synthesizer using the open tools HTS \cite{HTS} and Merlin \cite{Merlin}. First, we generated phoneme alignment on the corpus by using HTS. Next, we used Merlin to train the deep neural network-based synthesizer model consisting of acoustic and voice duration models. The linguistic features include 598 binary features and 44 continuous features. We also added a 6-dimensional emotion vector obtained from the corresponding speech sample to the input feature.

The synthesizer model output a 181-dimensional vector including static and dynamic speech parameters for the vocoder. Each acoustic and voice duration model was constructed by a feed-forward network with 6 hidden layers $\times$ 1024 units. We used $\tanh$ as an activation function. The training parameters were the same as the default ones used in an example of Merlin (\verb|egs/slt_arctic/s1|).

\section{Collection of Caregiver Speech}
\label{s:voice_analysis}
\subsection{Speech Collection Method}
\begin{table}[tb]
\caption{Profiles of caregivers whose speech was collected.}
	\label{table:careworkers}
	\centering
	{\footnotesize
		\begin{tabular}{lrrrr}
			\hline
			Item & Mean & S.D. & Min & Max \\ \hline 
			Age & 45.7 & 10.8 & 30 & 63 \\ 
			Years of Experience & 16.7 & 7.8 & 7 & 37 \\ 
			\hline
		\end{tabular}
	}
\end{table}

Twenty caregivers working in two residential elderly care facilities in Tsukuba City, Japan, participated in the collection, ten from each facility (8 female, 2 male). All caregivers had at least three years of experience. \tarf{table:careworkers} shows the profiles of the caregivers.

\begin{table*}[tb]
	\caption{Experimental situations described to participants in speech collection.}
	\label{table:situation}
	\centering
	{\footnotesize
		\begin{tabular}{lll}
			\hline
			Time of Day & Elderly Utterance & Given Situation \\\hline
			Early Morning & Opening Words & Mr./Ms. Sato woke up right now. \\
			& Positive Utterance & Mr./Ms. Sato said ``I'm feeling well." \\
			& Negative Utterance & Mr./Ms. Sato said ``I'm not feeling well." \\ \hline
			Forenoon & Opening Words & It's time to do some activities. But Mr./Ms. Sato doesn't seem to have any plans. \\
			& Positive Utterance & 	Mr./Ms. Sato said ``I'll take a walk." \\
			& Negative Utterance & Mr./Ms. Sato said ``I don't want to do anything today." \\ \hline
			Before Bedtime & Opening Words & It's time for Mr./Ms. Sato to go to bed. \\ 
			& Positive Utterance & Mr./Ms. Sato said ``I think I'll sleep better today." \\
			& Negative Utterance 1 & Mr./Ms. Sato said ``I don't think I can sleep today" and looks anxious. \\
			& Negative Utterance 2 & Mr./Ms. Sato said ``I don't want to go to bed yet" and looks energetic. \\ \hline
			Middle of the Night & Opening Words & 	Mr./Ms. Sato woke up in the middle of the night and couldn't seem to sleep. \\
			& Positive Utterance & Mr./Ms. Sato said ``I woke up to go to the restroom. I'm going back to sleep."
			 \\
			& Negative Utterance & Mr./Ms. Sato said ``I'm awake. I'm feeling restless." \\
			\hline
		\end{tabular}
	}
\end{table*}

The speech was collected in a meeting room at the caregivers' facilities. We began by describing a situation to a participant. The participant then walked from the starting point (assumed to be the entrance of the elderly person's room) and spoke to a doll which was used in place of an actual elderly person. To help them visualize the situation, we placed a photo of an elderly person close to the doll. We asked the participants to speak to the doll as they would to an elderly person.

The situations described to the participants are shown in \tarf{table:situation}. We assume the caregivers' speech can help regulate the listener's circadian rhythm by adjusting the emotional state of their voice depending on the time of day. Thus, we established four situations at different times throughout the day. In addition, we anticipated that the caregivers would change how they speak depending on the elderly person's utterance that they are responding to. Therefore, we created various conditions to account for the elderly's utterances. Specifically, for each time of day we created three or four situations: when the caregiver begins a conversation with the elderly person (opening words), when the elderly person uttered a sentence with positive sentiment (positive utterance), and when the elderly uttered a sentence with negative sentiment (negative utterance). A total of 13 situations were created, and they were described to the participants in the order shown in \tarf{table:situation}. A speech recording of a single time-of-day situation consisted of two sessions. In each session, a different elderly person was described to the caregiver in terms of gender, age, hearing ability, and cognitive ability. We randomly chose the status of the elderly person. The utterances were recorded with a headset microphone (SHURE SM10A-CN). The sampling rate for recording was 44.1kHz, and the bit depth was 24. From the recorded data, we extracted the speech duration when the participants actually spoke to the doll, resulting in 96 minutes of the caregivers' speech data.

\subsection{Results of Emotion Recognition}
\begin{figure}[tb]
	\centering
	\includegraphics[width=0.85\linewidth]{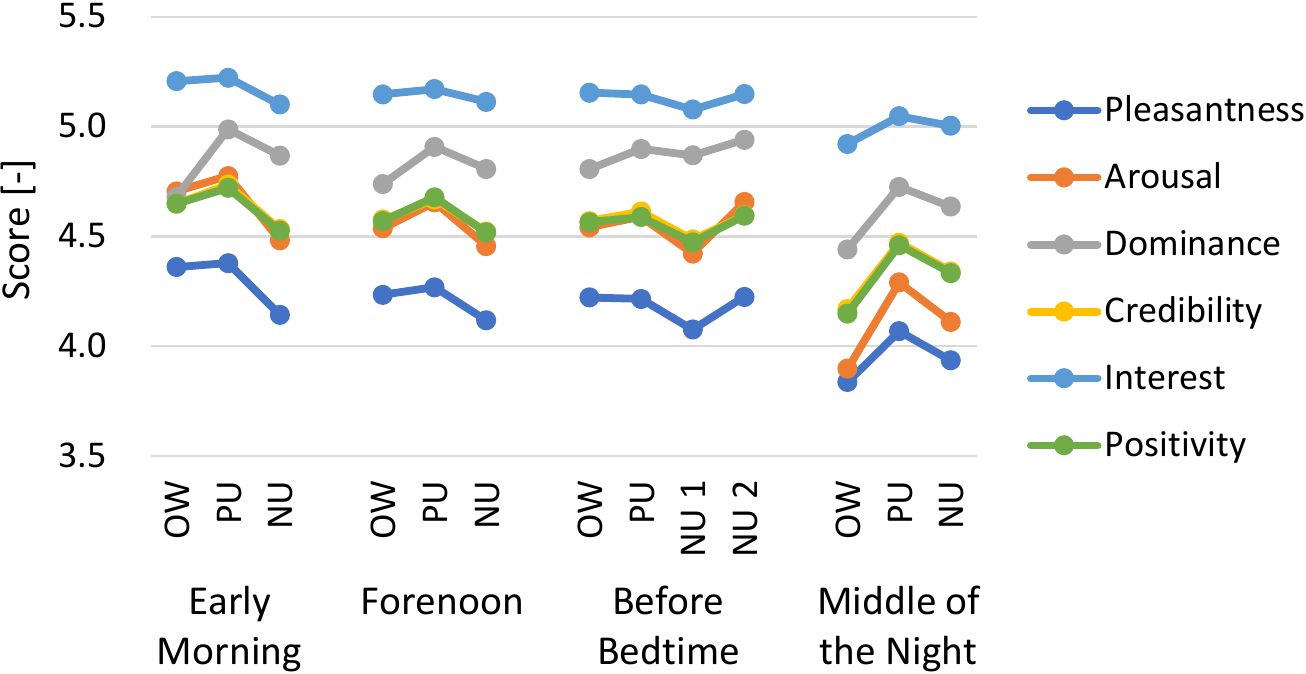}
	\caption{Emotion recognition results of caregivers' voices (OW: Opening Words, PU: Positive Utterance, NU: Negative Utterance).}
	\label{fig:obj}
\end{figure}

We obtained the emotion vectors of the caregivers' utterances by inputting them to the speech emotion recognizer. \firf{fig:obj} shows the average of the vectors in each situation. The emotion value of each dimension increases when the voice sounds ``stronger" and decreases when it sounds ``weaker." As shown in \firf{fig:obj}, we observed lower values in the middle of the night than in the other time-of-day situations, showing that most caregivers use ``weaker" voices when speaking to elderly people in the middle of the night. Looking at the difference among the elderly people's utterances, we observed higher values in positive utterances and lower values in negative utterances. The values in the opening words were in the middle of the two. The values in the negative utterances were likely lower because many caregivers speak with a somewhat soothing voice in response to the elderly person's negative utterances.

The changes in dimensions mainly correlated with each other. This was also observed by Mori et al.\cite{MoriAST}. Only the change in dominance was different from the others. In the opening words in the early morning, for instance, the dominance value was lower whereas the others were higher. We listened to the utterances in this situation and determined that the speech sounded somewhat cheerful but not overly so. Such detailed emotional expression could be reflected in the 6-dimensional emotion space.

\begin{table}[tb]
	\caption{Utterances in subjective evaluation.}
	\label{table:utterances}
	\centering
	{\footnotesize
		\begin{tabular}{ll}
			\hline
			Time of Day & Utterance \\
			\hline
			Early Morning & 
				Good morning, Mr./Ms. Sato. \\
				& Did you sleep well last night? \\ \hline
			Forenoon  & 
				Hello, Mr./Ms. Sato. It's a beautiful day today. \\
				& Would you like to take a walk? \\ \hline
			Before Bedtime & 
				Mr./Ms. Sato, it's time to go to bed. \\
				& Why don't you go to your room and sleep soon? \\ \hline
			Middle of the Night & 
				Mr./Ms. Sato, are you okay? Can't you sleep? \\ \hline
			
		\end{tabular}
	}
\end{table}

\section{Subjective Evaluation}
We synthesized speech samples on the basis of the emotional parameters observed in the caregivers' speech. We asked the elderly participants to listen to the samples spoken by a robot and investigated how the samples changed the participant's self-perceived arousal level, which is the most important metric for measuring the regulation of circadian rhythm.
To examine whether the speech generated by our method regulated the participants' rhythm, we first hypothesized that the voices with emotional control would affect the listener's self-perceived arousal level more than the voices without emotional control. The elderly participants were asked to compare two speech samples, with and without emotional control, and evaluate how their self-perceived arousal level changed.

\subsection{Method}
\subsubsection{Speech Generation}
\begin{figure}
	\centering
	\includegraphics[width=0.86\linewidth]{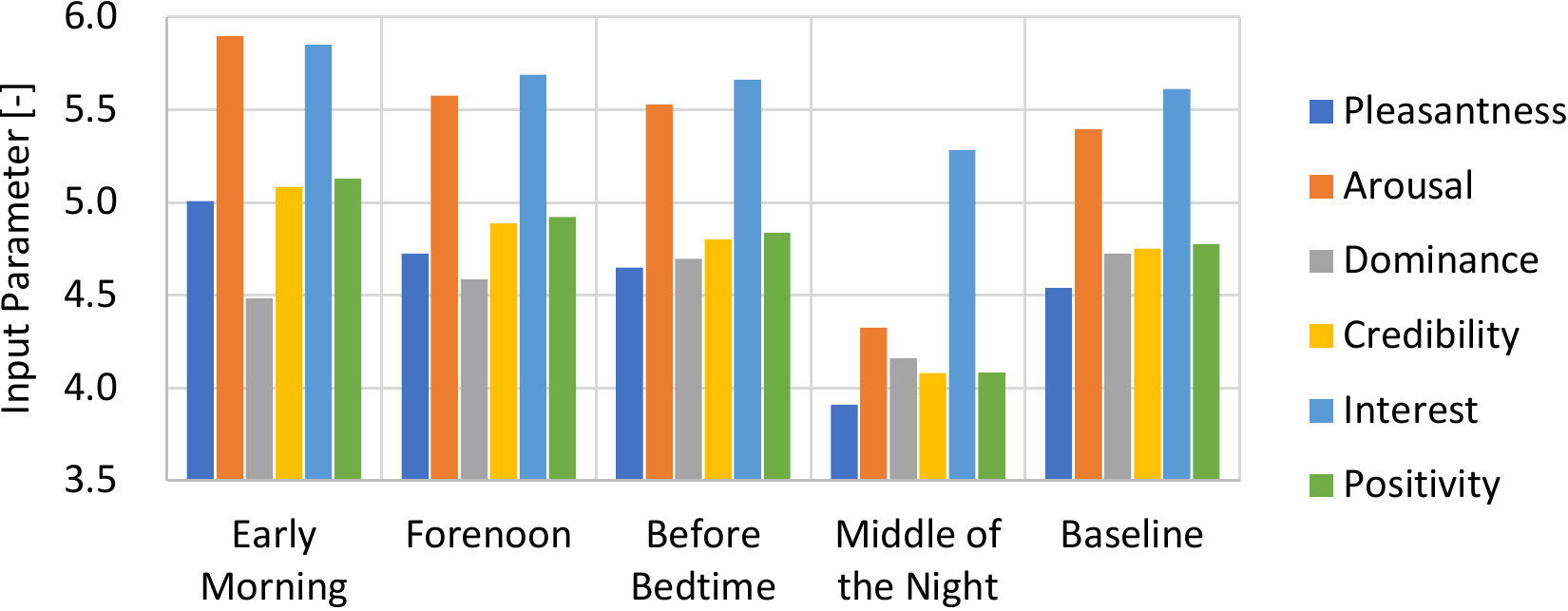}
	\caption{Input emotional parameters for speech synthesis.}
	\label{fig:syn_emotion}
\end{figure}

We first calculated the emotion vectors to be input to the synthesizer using \eqrf{eqn:conv}. We chose 15 target speakers from the 20 caregivers who participated in the speech collection. Because the voice actor for the speech synthesis corpus is female, we chose the female speakers for consistent voice quality between the corpus and target speech to maintain the quality of the synthesized voices. Out of 16 female speakers, one speaker's speech did not show significant emotional differences between the early morning and the middle of the night, so we excluded this speaker from the target speakers.

We only tested the utterances from the opening words for all time-of-day situations. The emotional parameters for synthesis were calculated on the basis of the target speech in each opening word situation. The adjustable coefficient $\alpha$ in \eqrf{eqn:conv} was set to 1.1. This value was used because, in our preliminary experiment, it was slightly difficult to distinguish the emotional difference between generated voices across different situations when $\alpha$ was 1, i.e., the as-is condition. Thus, we slightly emphasized the emotional difference by using a slightly larger value. \firf{fig:syn_emotion} shows the emotional parameters for speech synthesis. \tarf{table:utterances} shows the sentences spoken by the robot. These sentences were designed using phrases that appeared frequently in the caregivers' collected speech.

We refer to the condition where the speech emotion is controlled with our method as the proposed condition. For comparison, we created a baseline condition with speech samples without emotional control by using the same emotional parameters across all the situations. To calculate the emotional parameters for the baseline condition, we set $\alpha$ to 0 in \eqrf{eqn:conv}. The emotional parameters for the baseline are shown in \firf{fig:syn_emotion}.

\begin{figure}
	\centering
	\includegraphics[width=0.25\linewidth]{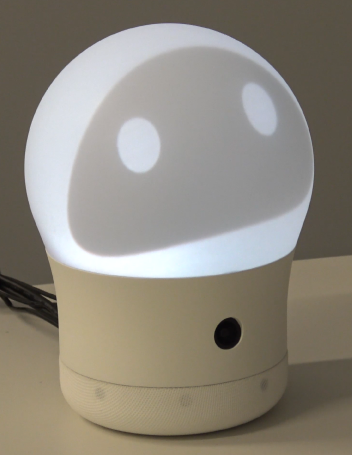}
	\caption{Experimental robot.}
	\label{fig:magnus}
\end{figure}

\subsubsection{Subjective Evaluation}
\firf{fig:magnus} shows the robot that was used in the experiment. The robot is shaped like a light bulb and shows facial expressions through eye movements. The robot always nodded twice as it started to speak.

Before the evaluation, each participant had a conversation session where the participant talked with the robot through four different scenarios simulating the different time-of-day situations\footnote{ For instance, in the early morning scenario, the robot initiates a dialogue by saying "Good morning, Mr./Ms. Sato. Did you sleep well last night?" If the participant responds with a positive utterance, the robot says ``I'm glad to hear that you're feeling well." If the participant responds with a negative utterance, the robot says ``I see that you are not feeling so well. Do you have any pain or do you feel sick?" }. Each scenario consisted of two dialogues. One dialogue contained at least two conversational turns. After the session, the participants started the evaluation session, where they were asked to compare two speech samples (baseline and proposed) played from the robot in a sequence.

\tarf{table:subj_axis} shows the metrics of the comparison. For each metric, the participant rated which speech sample was higher using the Visual Analogue Scale (VAS). The order in which the speech samples (baseline or proposed) was played was counter-balanced across the participants.

The participants were members of Center for Usability and Aging Research at the University of Tsukuba \cite{Minlab}. There were 20 participants in total (10 female, 10 male), all over 70 years old (average 74.60, S.D. 2.28). They did not have any cognitive (MMSE score over 26) or hearing impairments.

\begin{table}[tb]
	\caption{Evaluation metrics of subjective evaluation.}
	\label{table:subj_axis}
	\centering
	{\footnotesize
		\begin{tabular}{ll}
			\hline
			Item & Direction \\ \hline
			Arousal & Which voice made you feel more active? \\
			Valence & Which voice made you feel more comfortable? \\
			Naturalness & Which voice sounded more natural? \\
			Preference & Which voice did you prefer? \\
			\hline
		\end{tabular}
	}
\end{table}

\subsection{Result}
\label{ss:subj_eval_elderly}
\begin{figure}[tb]
	\centering
	\includegraphics[width=0.99\linewidth]{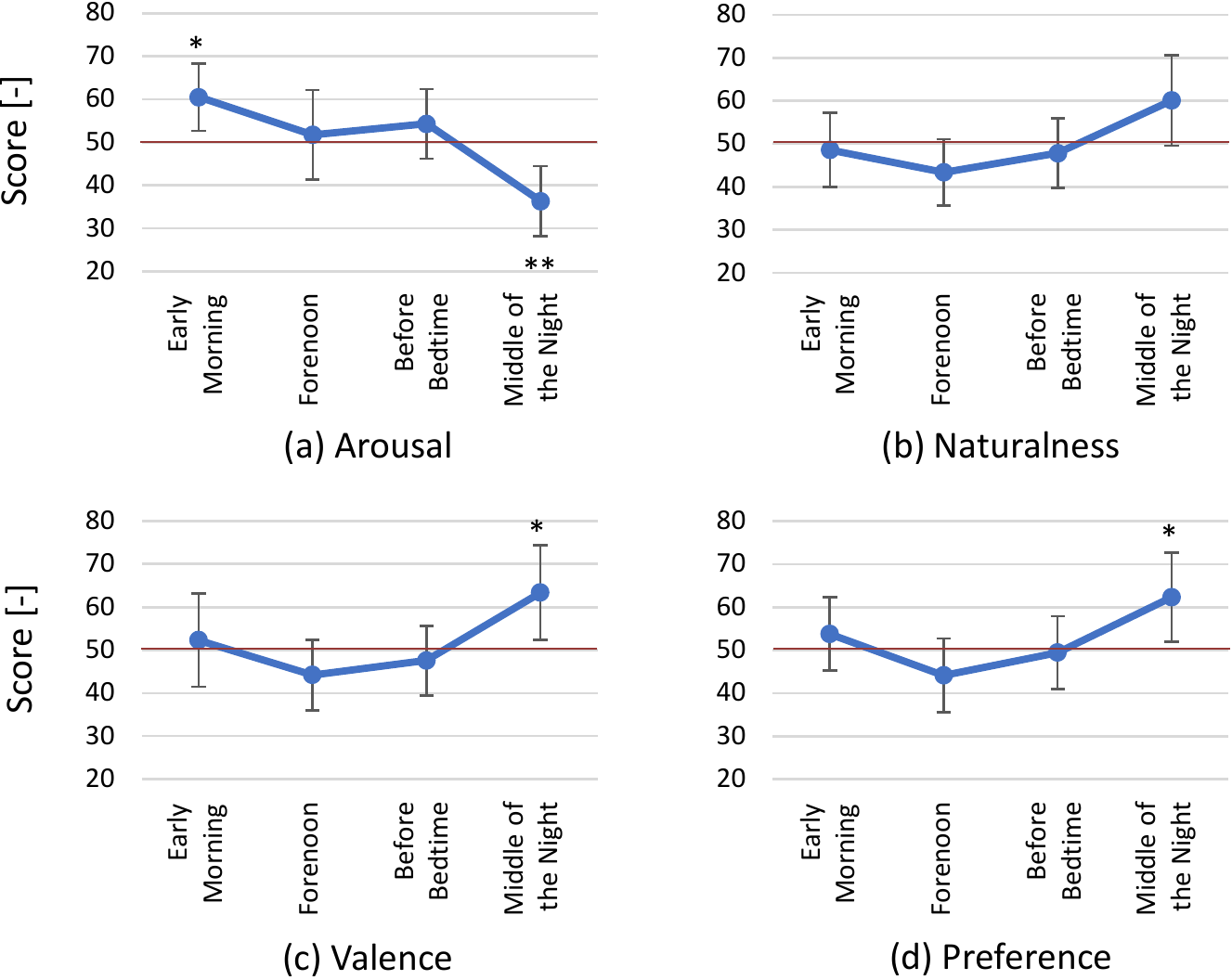}
	\caption{Subjective evaluation results of emotional voices and baselines. Error bars indicate 95\% confidence intervals. Asterisks indicate significant differences from 50, i.e., the tie-judgment level (One sample $t$-test with two-tailed hypothesis; *$p<0.05$, **$p<0.01$).}
	\label{fig:direct_comparison}
\end{figure}

VAS is converted to a continuous score ranging from 0 to 100. The converted score is higher than 50 if the participant rated the proposed speech sample higher than the baseline sample. \firf{fig:direct_comparison} shows the scores of the evaluation averaged across all the participants. If a pair of speech samples (baseline and proposed) are judged as equivalent, the score is 50, which is referred to as the ``tie-judgment level." \firf{fig:direct_comparison} also shows asterisk marks in the conditions where we observed significant differences from the tie-judgment level.

The arousal level (a) was high in the early morning and low in the middle of the night. The results of both situations showed significant differences from the tie-judgment level. This indicates that listening to speech samples generated by the proposed method made the participants feel more active in the early morning and calmer in the middle of the night, which supports our hypothesis.

The naturalness (b) showed no significant difference from the tie-judgment level in all situations. This result indicates that the manipulation of speech emotion did not cause severe degradation in the naturalness of the synthesized voices.

For valence (c) and preference (d), we only observed a significant difference from the tie-judgment level in the middle of the night, which resulted in a higher score for the proposed condition than the baseline. As shown in \firf{fig:syn_emotion}, the emotional parameters in the middle of the night differed greatly from that of the other situations. This result indicates that the emotions of the robot's voices in the middle of the night were in line with what the participants expected.

\subsection{Discussion}
The subjective evaluation showed that the speech synthesizer with our proposed method successfully produced a robot voice that influenced the participant's self-perceived arousal level. From this result, we conclude that our speech synthesis method has the possibility to enable robots to influence people through emotional speech similarly to people.

Unlike the previous emotional speech synthesis methods, our method requires less effort on the developer's part to adjust the emotional parameters of the synthesized voices. In our method, the developers must adjust only one parameter, the coefficient $\alpha$ in \eqrf{eqn:conv}. Although the manual adjustment was minimized, we could obtain synthesized speech with enough emotion to adjust the elderly participants' self-perceived arousal level. Thus, we verified that reducing the need to manually adjust the parameters is useful for practical application.

In the subjective evaluation, most participants preferred the proposed speech to the baseline ones. However, we also observed individual differences in their voice preferences. Four participants answered they preferred the baseline speech over the proposed in the middle of the night. In regards to this, the participants' expressed the following opinions:

\begin{itemize}
\item The participants thought the proposed speech in the middle of the night sounded as if ``the robot was mumbling in the dark" and made them feel depressed.
\item The participants felt that the baseline speech was more thoughtful and kind because they liked that the robot spoke in a consistent tone at all times of day.
\end{itemize}
To develop a robot that is suitable for many people, the above opinions suggest that it is necessary to customize the robot's speaking style depending on the user's individual preference.

\section{Conclusion}
Toward the development of a robot that can influence people through speech communication as humans do, we proposed a speech synthesis method for imitating the speech emotions of human speakers. The subjective evaluation showed that the speech samples generated by our method could regulate the self-perceived arousal level of the elderly participants, indicating the possibility that our method can be used to develop a robot that can regulate the circadian rhythm of the elderly to maintain their health and improve their quality of life.

\bibliographystyle{IEEEtran} 
\bibliography{references} 

\begin{thebibliography}{10}
\providecommand{\url}[1]{#1}
\csname url@samestyle\endcsname
\providecommand{\newblock}{\relax}
\providecommand{\bibinfo}[2]{#2}
\providecommand{\BIBentrySTDinterwordspacing}{\spaceskip=0pt\relax}
\providecommand{\BIBentryALTinterwordstretchfactor}{4}
\providecommand{\BIBentryALTinterwordspacing}{\spaceskip=\fontdimen2\font plus
\BIBentryALTinterwordstretchfactor\fontdimen3\font minus
  \fontdimen4\font\relax}
\providecommand{\BIBforeignlanguage}[2]{{%
\expandafter\ifx\csname l@#1\endcsname\relax
\typeout{** WARNING: IEEEtran.bst: No hyphenation pattern has been}%
\typeout{** loaded for the language `#1'. Using the pattern for}%
\typeout{** the default language instead.}%
\else
\language=\csname l@#1\endcsname
\fi
#2}}
\providecommand{\BIBdecl}{\relax}
\BIBdecl

\bibitem{Inoue}
J.~Mizuno, D.~Saito, K.~Sadohara, M.~Nihei, S.~Ohnaka, J.~Suzurikawa, and
  T.~Inoue, ``Effect of the information support robot on the daily activity of
  older people living alone in actual living environment,'' \emph{International
  Journal of Environmental Research and Public Health}, vol.~18, 2021, article.
  2498.

\bibitem{IR}
``Intuition {Robotics} --- {Digital} companion technology,''
  \url{https://www.intuitionrobotics.com/}, accessed: 2021-02-25.

\bibitem{Yamagishi}
J.~Yamagishi, K.~Onishi, T.~Masuko, and T.~Kobayashi, ``Acoustic modeling of
  speaking styles and emotional expressions in hmm-based speech synthesis,''
  \emph{IEICE Transactions on Information and Systems}, vol.~88, no.~3, pp.
  502--509, 2005.

\bibitem{Lorenzo-Trueba}
J.~Lorenzo-Trueba, R.~Barra-Chicote, R.~San-Segundo, J.~Ferreiros,
  J.~Yamagishi, and J.~M. Montero, ``Emotion transplantation through adaptation
  in hmm-based speech synthesis,'' \emph{Computer Speech and Language},
  vol.~34, no.~1, pp. 292--307, 2015.

\bibitem{Xue}
Y.~Xue, Y.~Hamada, and M.~Akagi, ``Voice conversion for emotional speech:
  Rule-based synthesis with degree of emotion controllable in dimensional
  space,'' \emph{Speech Communication}, vol. 102, pp. 54--67, 2018.

\bibitem{Rabiee}
A.~Rabiee, T.-H. Kim, and S.-Y. Lee, ``Adjusting pleasure-arousal-dominance for
  continuous emotional text-tospeech synthesizer,'' in \emph{Proc.
  {INTERSPEECH} (Show and Tell)}, 2019, pp. 3693--3694.

\bibitem{Zhu}
X.~Zhu, S.~Yang, G.~Yang, and L.~Xie, ``Controlling emotion strength with
  relative attribute for end-to-end speech synthesis,'' in \emph{Proc. IEEE
  ASRU}, 2019, pp. 192--199.

\bibitem{Lorenzo-Trueba2}
J.~Lorenzo-Trueba, G.~E. Henter, S.~Takaki, J.~Yamagishi, Y.~Morino, and
  Y.~Ochiai, ``Investigating different representations for modeling and
  controlling multiple emotions in {DNN}-based speech synthesis,'' \emph{Speech
  Communication}, vol.~99, pp. 135--143, 2018.

\bibitem{Nose}
T.~Nose and T.~Kobayashi, ``Perceptual expressivity modeling technique for
  speech synthesis based on multiple-regression {HSMM},'' in \emph{Proc.
  {INTERSPEECH}}, 2011, pp. 109--112.

\bibitem{Hodari}
Z.~Hodari, O.~Watts, S.~Ronanki, and S.~King, ``Learning interpretable control
  dimensions for speech synthesis by using external data,'' in \emph{Proc.
  {INTERSPEECH}}, 2018, pp. 32--36.

\bibitem{Cai}
X.~Cai, D.~Dai, X.~Wu, Zhiyong~Li, J.~Li, and H.~Meng, ``Emotion controllable
  speech synthesis using emotion-unlabeled dataset with the assistance of
  cross-domain speech emotion recognition,'' in \emph{Proc. {IEEE} {ICASSP}},
  2021, pp. 5734--5738.

\bibitem{Hood}
S.~Hood and S.~Amir, ``The aging clock: Circadian rhythms and later life,''
  \emph{The Journal of Clinical Investigation}, vol. 127, no.~2, pp. 437--446,
  2017.

\bibitem{Livingston}
G.~Livingston, B.~Blizard, and A.~Mann, ``Does sleep disturbance predict
  depression in elderly people? {A} study in inner {London},'' \emph{British
  Journal of General Practice}, vol.~43, pp. 445--448, 1993.

\bibitem{Shankar}
R.~Shankar, J.~Sager, and A.~Venkataraman, ``A multi-speaker emotion morphing
  model using highway networks and maximum likelihood objective,'' in
  \emph{Proc. {INTERSPEECH}}, 2019, pp. 2848--2852.

\bibitem{Zhou}
K.~Zhou, B.~Sisman, M.~Zhang, and H.~Li, ``Converting anyone's emotion: Towards
  speaker-independent emotional voice conversion,'' in \emph{Proc.
  {INTERSPEECH}}, 2020, pp. 3416--3420.

\bibitem{Li}
T.~Li, S.~Yang, L.~Xue, and L.~Xie, ``Controllable emotion transfer for
  end-to-end speech synthesis,'' in \emph{Proc. International Symposium on
  Chinese Spoken Language Processing ({ISCSLP})}, 2021.

\bibitem{James}
J.~James, C.~I. Watson, and B.~MacDonald, ``Artificial empathy in social
  robots: An analysis of emotions in speech,'' in \emph{Proc. IEEE RO-MAN},
  2018, pp. 632--637.

\bibitem{Dmello}
S.~D'mello and A.~Grasser, ``Autotutor and affective autotutor: Learning by
  talking with cognitively and emotionally intelligent computers that talk
  back,'' \emph{ACM Transactions on Interactive Intelligent Systems}, vol.~2,
  no.~4, 2012.

\bibitem{Chiba}
Y.~Chiba, T.~Nose, M.~Yamanaka, T.~Kase, and A.~Ito, ``An analysis of the
  effect of emotional speech synthesis on non-task-oriented dialogue system,''
  in \emph{Proc. SIGDIAL}, 2018, pp. 371--375.

\bibitem{Nass}
C.~Nass, I.-M. Jonsson, H.~Harris, B.~Reaves, J.~Endo, S.~Brave, and
  L.~Takayama, ``Improving automotive safety by pairing driver emotion and car
  voice emotion,'' in \emph{Proc. CHI (Extended Abstracts)}, 2005, pp.
  1973--1976.

\bibitem{Karadogan}
S.~G. Karado\u{g}an and J.~Larsen, ``Combining semantic and acoustic features
  for valence and arousal recognition in speech,'' in \emph{Proc. International
  Workshop on Cognitive Information Processing (CIP)}, 2012.

\bibitem{Eyben}
F.~Eyben, M.~W\"{o}llmer, A.~Graves, B.~Schuller, E.~Douglas-Cowie, and
  R.~Cowie, ``On-line emotion recognition in a 3-d activation-valence-time
  continuum using acoustic and linguistic cues,'' \emph{Journal on Multimodal
  User Interfaces}, vol.~3, pp. 7--19, 2010.

\bibitem{Yang}
Z.~Yang and J.~Hirschberg, ``Predicting arousal and valence from waveforms and
  spectrograms using deep neural networks,'' in \emph{Proc. {INTERSPEECH}},
  2018, pp. 3092--3096.

\bibitem{Schmitt}
M.~Schmitt and B.~Schuller, ``Deep recurrent neural networks for emotion
  recognition in speech,'' in \emph{Proc. Jahrestagung f\"ur Akustik ({DAGA})},
  March 2018, pp. 1537--1540.

\bibitem{Atmaja}
B.~T. Atmaja and M.~Akagi, ``Dimensional speech emotion recognition from speech
  features and word embeddings by using multitask learning,'' \emph{{APSIPA}
  Transactions on Signal and Information Processing}, vol.~9, p. e17, 2020.

\bibitem{Grimm}
M.~Grimm, K.~Kroschel, E.~Mower, and S.~Narayanan, ``Primitives-based
  evaluation and estimation of emotions in speech,'' \emph{Speech
  Communication}, vol.~49, no. 10-11, pp. 787--800, 2007.

\bibitem{Parthasarathy}
S.~Parthasarathy and C.~Busso, ``Jointly predicting arousal, valence and
  dominance with multi-task learning,'' in \emph{Proc. {INTERSPEECH}}, 2017,
  pp. 1103--1107.

\bibitem{Moore}
J.~D. Moore, L.~Tian, and C.~Lai, ``Word-level emotion recognition using
  high-level features,'' in \emph{Proc. International Conference on
  Computational Linguistics and Intelligent Text Processing ({CICLing})}, 2014,
  pp. 17--31.

\bibitem{Schuller}
B.~Schuller, M.~Valster, F.~Eyben, R.~Cowie, and M.~Pantic, ``{AVEC} 2012: The
  continuous audio/visual emotion challenge,'' in \emph{Proc. ACM ICMI}, 2012,
  pp. 449--456.

\bibitem{MoriAST}
H.~Mori, H.~Kasuya, M.~Nakamura, and M.~Amanuma, ``Some considerations for
  designing spoken dialogue database from the viewpoint of paralinguistic
  information,'' \emph{Acoustical Science and Technology}, vol.~24, no.~6, pp.
  376--378, 2003.

\bibitem{1400Mori}
H.~Mori, A.~Nagaoka, and Y.~Arimoto, ``Accuracy of automatic cross-corpus
  emotion labeling for conversational speech corpus commonization,'' in
  \emph{Proc. LREC}, 2016, pp. 4019--4023.

\bibitem{Fujioka}
T.~Fujioka, T.~Homma, and K.~Nagamatsu, ``Meta-learning for speech emotion
  recognition considering ambiguity of emotion labels,'' in \emph{Proc.
  {INTERSPEECH}}, 2020, pp. 2332--2336.

\bibitem{JEITAKikaku}
{Japan Electronics and Information Technology Industries Association},
  ``{IT-4012}: Guidelines for speech types to express emotions and
  intentions,'' 2018, (in Japanese).

\bibitem{HTS}
``{HMM/DNN}-based speech synthesis system ({HTS}),''
  \url{http://hts.sp.nitech.ac.jp/}, accessed: 2021-03-09.

\bibitem{Merlin}
``Merlin,'' \url{http://www.cstr.ed.ac.uk/projects/merlin/}, accessed:
  2021-03-09.

\bibitem{Minlab}
E.~T. Harada, ``Taking off! {Founding} the center for usability and aging
  research ({CUAR}) project,'' \emph{Journal of Human Life Engineering},
  vol.~13, no.~1, pp. 71--74, 2012, (in Japanese).

\end{thebibliography}

\end{document}